\documentclass[a4paper,10pt,twoside]{article}

\usepackage{clin}        % Stylefile for CLIN Journal
\usepackage{harvard}     % Bibliography Stylefile
\usepackage{url,mdwlist,booktabs,color}
\usepackage[pdftex,pdfborder={0 0 0},linkcolor=red,urlcolor=blue,colorlinks,unicode,breaklinks,hyperfootnotes=false]{hyperref}

\begin{document}

\title{The Hebrew Bible as Data: Laboratory – Sharing - Experiences}

\author{Dirk Roorda$^*$ $^{**}$ \email{dirk.roorda@dans.knaw.nl}
\AND \addr{$^*$Data Archiving and Networked Services - Royal Netherlands Academy of Arts and Sciences, Anna van Saksenlaan 10; 2593 HT Den Haag, Netherlands}
\AND \addr{$^{*}$The Language Archive - Max Planck Institute for Psycholinguistics, Wundtlaan 1; 6525 XD Nijmegen, Netherlands}
}

\maketitle\thispagestyle{empty}

\begin{abstract}
The systematic study of ancient texts including their production, transmission and interpretation is greatly aided
by the digital methods that started taking off in the 1970s.
But how is that research in turn transmitted to new generations of researchers?
We tell a story of Bible and computer across the decades and then point out the current challenges:
(1) finding a stable data representation for changing methods of computation;
(2) sharing results in inter- and intra-disciplinary ways, for reproducibility and cross-fertilization.
We report recent developments in meeting these challenges.
The scene is the text database of the Hebrew Bible, constructed by the Eep Talstra Centre for Bible and Computer (ETCBC),
which is still growing in detail and sophistication.
We show how a subtle mix of computational ingredients enable scholars to
research the transmission and interpretation of the Hebrew Bible in new ways:
(1) a standard data format, Linguistic Annotation Framework (LAF);
(2) the methods of scientific computing, made accessible by (interactive) Python and its associated ecosystem.
Additionally, we show how these efforts have culminated in the construction of a new,
publicly accessible search engine SHEBANQ, where the text of the Hebrew Bible
and its underlying data can be queried in a simple, yet powerful query language MQL,
and where those queries can be saved and shared. 
\end{abstract}

\section{Introduction}
The Hebrew Bible is a collection of ancient texts resulting from a ten-centuries long tradition.
It is one of the most studied texts in human culture.
Information processing by machines is less than two centuries old, but since its inception its capabilities
have evolved in an exponential manner up till now \cite{Gleick2011}.
We are interested in what happens when the Hebrew Bible as an object of study is brought
under the scope of the current methods of information processing.
The Eep Talstra Centre for Bible and Computing (ETCBC) formerly known as Werkgroep Informatica Vrije Universiteit (WIVU),
has been involved in just this since the 1970s and their members are dedicated to this approach.
The combination of a relatively stable set of data and a rapidly evolving set of methods urges for reflection.
Add to that a growing set of ambitious research questions,
and it becomes clear that not only reflection is needed but also action.
Methods from computational linguistics and the wider digital humanities are to be used,
hence people from different disciplines have to be involved. How can the ETCBC share its data
and way of working productively with people that are used to a wide variety of computational ways?

In this article we tell a story of reflection and action,
and the characters are databases, data formats, query languages, annotations,
computer languages, archives, repositories and social media.
This story has a beginning in February 2012, when a group of biblical scholars convened
at the Lorentz center at Leiden for the workshop
Biblical Scholarship and Humanities Computing: Data Types, Text, Language and Interpretation \cite{Roorda2012c}.
They searched for new ways to obtain computational tools that matched their research interests.
The author was part of that meeting and had prepared a demo application: a query saver.
It was an attempt to improve the sharing of knowledge.
It is a craft to write successful queries for the ETCBC Hebrew Text database, and by publishing their queries,
researchers might teach each other how to do it.

In the years that followed, this idea has materialized as the result of the
SHEBANQ project (System for HEBrew text: ANnotations for Queries and markup),
a curation and demonstrator project funded by CLARIN-NL,
the Dutch department of the Common LAnguage Resource INfrastructure in Europe \url{http://www.clarin.eu}.
We have chosen a modern standard format for the data: Linguistic Annotation Framework (LAF),
and have built a web-application for saving queries.
During the execution of this project we also have built LAF-Fabric, a tool to analyze and manipulate LAF resources.
Now, in 2014, we can say that we have a modern data laboratory for historico-linguistic data, plus ways to share results,
not only among a small circle of theological experts,
but also among computational linguists on the one hand and students and interested lay people on the other.

Of course, every beginning of such a story is arbitrary.
There is always so much more that happened before.
In order to provide the reader with enough context, we shall also relate key moments of that greater story.
Moreover, we cannot tell the whole story: our perspective is biased to the computational side.
We shall not delve into the intricacies of manuscript research,
but focus on the data models and computational methods that help analyze a rather fixed body of transcribed text.
Yet, we believe that this simplified context is rich enough material for a good story.
Whereas this paper deliberately scratches only the surface of the computational methods,
there is also a joint paper with researchers, which contains a more technical account in \cite{Roorda2014a}.

\section{Ground work: WIVU and ETCBC}
Since the 1970s, Eep Talstra, Constantijn Sikkel and a group of researchers at the VU University Amsterdam
have been compiling a text database of the Hebrew Bible.
This database started as a set of files, containing the transliterated Hebrew text of the Bible
according to the Biblia Hebraica Stuttgartensia edition \cite{BHS}.
To this text, they added files with their observations of linguistic patterns in it as coded annotations,
anchored to the individual words, phrases, clauses, and sentences.
They tested tentative patterns against the data, refined them, and added manual exceptions.
This led to a complex web of files, containing the base text and a set of semi-automatically generated annotations.
They refrained from shaping these annotations in a hierarchical, linguistic model,
because they wanted to represent observations, not theory \cite{Talstra2000}.
The result of this work is a database in the sense of being observational data on which theories can be based.
It is not a database in the sense of a modern relational database system. 

The advantages of a proper (in the sense of computer science) database are obvious indeed,
but the relational model does not represent textual data in a natural way,
and does not facilitate queries that are linguistically meaningful.
In the 1990s there have been promising efforts to define the notion of a text database.
In his Ph.D. thesis, Crist-Jan Doedens \cite{Doedens1994} defined a data model for texts and
the notion of a topographic query language (QL) to retrieve linguistic results.
He identified the relations of sequence and embedding as the key structures to store and retrieve texts.
A query is topographic if its internal structure exhibits the same
sequence and embedding relations as the results it is meant to retrieve.
Interestingly, he did not postulate that a text is one hierarchy.
In his data model, textual data may be organized by means of multiple, overlapping hierarchies.

The definition of a data model and a query language are not yet a working database system.
In the 2000s, Ulrik Petersen undertook to create an implementation of Doedens’s ideas.
This led to the Emdros database system with the MQL (Mini-QL) query language \citename{Petersen2004}~\citeyear{Petersen2004,Petersen2006,Petersen2002}.
Emdros consists of a front-end, which is an MQL interpreter, and a back-end, which is an existing
production class relational database system such as Postgres or MySQL.
Despite the fact that MQL is a concession to practicality, it is still a topographic query language
and very convenient to express real-life textual queries without invoking programming skills.
Since then, an Emdros export of the current Hebrew text database is being maintained by the ETCBC team.
Emdros is open source software, the data model is very clear, so this export is a {\itshape communication device}:
the intricacies of the internal annotation-creation of the ETCBC workflow are largely left behind,
and users of the export have a well-defined dataset at their disposal.

\section{Idea: Queries As Annotations}
During the aforementioned Lorentz workshop \cite{Roorda2012c},
an international group of experts reflected on how to bring biblical data resources to better fruition in the digital age.
The ETCBC database had been incorporated in Bible study software,
but developments there were not being driven by agendas set by academic research.
Yet those bible study applications offered attractive interfaces to browse the text, look up words and more.
The problem was: how can theologians, with limited ICT resources,
regain control over the development of software that works with their data?
The workshop offered no concrete solutions, but some ingredients of potential long-term solutions did get mentioned:
open up the data and develop open source tools.
Theologians can only hope to keep up with ICT developments if they allow people building on each others’ accomplishments.
A very concrete articulation of this statement was made by Eep Talstra himself,
when he deposited the ETCBC database into EASY, the research archive of DANS \cite{Talstra2012}.
It must be admitted that there remained barriers:
the data was not Open Access and the format in which it was deposited was MQL,
which is not a very well-known format, so the experimenting theological programmer
still has a hard time to do some meaningful work with this data.
But it was definitely a step towards increased sharing of resources. 

\begin{figure}[htb]
\includegraphics[scale=0.9]{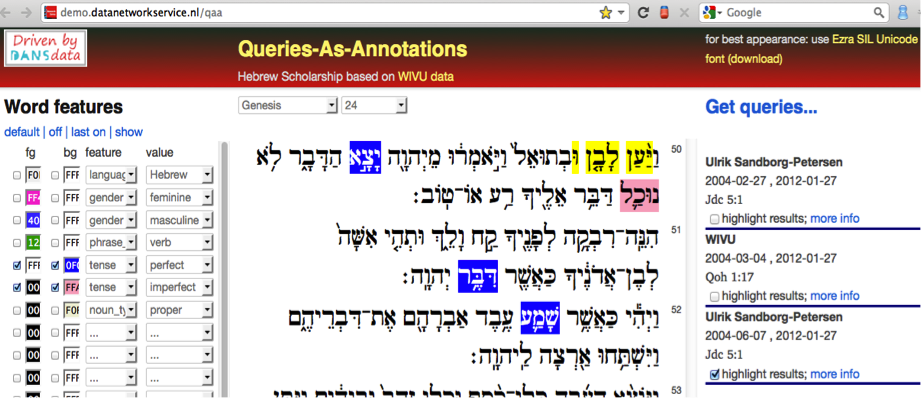}
\caption{Queries/Features as Annotations}
\label{qaa}
\end{figure}

In that same workshop, the author showed a demo application \cite{Roorda2012b} (see Figure~\ref{qaa})
by which the user could browse the Hebrew text and highlight a number of linguistic features.
The idea to highlight features, which are essentially annotations to the text, triggered another idea:
to view queries as annotations to the passages that contain their results \cite{Roorda2012a}.
If researchers can save their carefully crafted queries as annotations, and if those annotations are centrally stored,
then other researchers have access to them and may encounter them when they are reading a passage.
Just as readers encounter ordinary annotations by other scholars in printed books, they will encounter
results of queries of others when they are browsing a chapter of the Hebrew Bible in their web browser.
With a single click they are led to not only the query instruction itself but also a description of
the provenance and motivation of the query.
This could be the basis of interesting scenarios for cross-fertilization. 

It is interesting to note the stack of computational tools needed to write this demo.
Its construction involved a data preparation tool for transforming the contents of the ETCBC database
into a relational database for driving a website.
The web app itself was based on web2py, a lightweight python based web-application framework \cite{Pierro}. 

Table~\ref{softstack} is a list of languages used to implement both the data-preparation tool and the web-site,
together with the amount of code needed in each formalism. There are several things to note: 

\begin{enumerate}
\item The numbers of lines of code are very small.
\item The formalisms, while considerable in number, are utterly commonplace. 
\item The number of formalisms may be reduced by one by dropping Perl in favor of Python
\end{enumerate}

It can be concluded that mastering commonplace ICT techniques may generate a good return on investment,
in the form of a web application that expose data on the web in rich interfaces. 

\begin{table}[htb]
	\centering
	\begin{tabular}{lrr}
	\toprule
{\small\bfseries formalism}&{\small\bfseries	web app}&{\small\bfseries data prep tool}\\
sql&90&80\\
python&250&\\
perl&&650\\
javascript&300&\\
html&50&\\
css&60&\\
shell script&&280\\
	\bottomrule
	\end{tabular}
	\caption{Amount of lines of code per formalism per application}
	\label{softstack}
\end{table}

\section{Realization: LAF-Fabric and SHEBANQ}
In 2013-2014, ETCBC together with DANS has carried out the CLARIN-NL project SHEBANQ.
We seized the opportunity to implement the idea of queries-as-annotations,
but to make it possible at all more work had to be done. 

\subsection{LAF and LAF-Fabric}
First of all, a new representation of the data had to be selected, one that conformed to a standard used in linguistics.
Linguistic Annotation Framework, an ISO standard \cite{Ide2012}, was chosen.
LAF defines a data model in which an immutable stream of primary data is annotated by feature structures.
The data stream is addressed by means of a graph of nodes and edges, where the nodes may be linked to regions
of the primary data, and where edges serve to connect smaller parts to bigger wholes.
Both nodes and edges can act as targets of annotations, which contain the feature structures.
Finally, all entities, except the primary data, are serialized in XML.

In concrete terms, we have extracted the complete text of the Hebrew Bible as a plain Unicode text file.
As far as LAF is concerned, this is our primary data.
For the books, chapters and verses we have created nodes that are linked to the stretches of text that they correspond to.
For every individual word there is a node, linked to a region defined by the character positions
of the first and last character of that word.
For the phrases, clauses and sentences there are nodes, linked to the regions corresponding to the words they contain.
Relationships between constituents correspond to edges.
The properties of sectional units, words, and constituents are key-value pairs targeted at the corresponding nodes.

The LAF data model shares a lot of structure with the Emdros data model of text, objects and features.
We only had to map objects to nodes and features to key-value pairs inside annotations targeting the proper nodes,
so this conversion has been a straightforward process with only a few devilish details.

The result is a good example of stand-off markup.
The primary data is left untouched, and around it is a graph of annotations.
It is perfectly possible to add new annotations without interfering with the primary data or the other annotations.
The annotations are like a fabric, into which new threads can be woven, and that can be stitched to other fabrics.
In this way, the stand-off way of adding information to sources facilitates cooperation and sharing much better
than adding markup inline, such as TEI prescribes.
This bold assertion must be qualified by two considerations, however:

\begin{enumerate}
\item Stand-off markup works best in those cases where the primary sources are immutable.
As easy as it is to add new annotations, so difficult it is to insert new primary data.
\item Stand-off markup flourishes in cases where the main access mode to the sources is by programmatic means.
Manual inspection of stand-off data and their annotations becomes quickly overwhelming.
\end{enumerate}

In our case, condition 1 is satisfied for years in a row.
How we will deal with major updates remains to be seen.

Table~\ref{quantities} indicates some quantities of the ETCBC data, both in their Emdros form and in their LAF form.
These numbers suggest that manual inspection of individual files is so cumbersome
that it pays off to invest in programmatic access of the data.

\begin{table}[htb]
	\centering
	\begin{tabular}{lrr}
	\toprule
{\small\bfseries quantity}&{\small\bfseries	Emdros}&{\small\bfseries LAF}\\
words&426,555&426,555\\
linguistic objects resp. nodes&945,726&945,726\\
total number of features&22,622,100&25,504,388\\
serialized size (MQL resp. XML)&455 MB in 1 file&1640 MB in 14 files\\
compiled size (SQLite3 resp. binary)&126 MB&260 MB\\
	\bottomrule
	\end{tabular}
	\caption{Quantities in the ETCBC data}
	\label{quantities}
\end{table}

The LAF version of the Hebrew text database has been archived at Data Archiving and Networked Services (DANS),
the research archive for the humanities and social sciences in the Netherlands \cite{Peursen2014a}.

As LAF is a relative new standard, there are few LAF-compatible tools.
A LAF resource is represented in XML,
but the nature and size of this XML make it difficult to be handled by ordinary XML tools.
Looking through the surface syntax, a LAF resource is neither a relational database, nor a document, but a graph.
XML processing works well when the underlying data structure is a single hierarchy,
no matter how deep, or a table of records, no matter how large,
but it grinds to a halt when the data is a large and intricate web of nodes and edges, i.e. a graph.

In order to facilitate productive work with the freshly created LAF representation of the Hebrew Bible,
we have developed LAF-Fabric \cite{Roorda2013}, which is a LAF compiler and loader.
In a typical workflow, a researcher wants to inspect the LAF data, focus on some aspects,
sort, collate, link and transform selected data, and finally export results.
Without LAF-Fabric, the obvious way to do so is read the XML data, apply XPATH, XSLT or XQUERY scripts
and collect the results.
Reading the XML data means parsing it and building an internal representation in memory,
and this alone takes an annoying 15 minutes on a average laptop and uses a prohibitive amount of memory.
This is not conducive to an interactive, explorative, agile use of the data, and LAF-Fabric remedies this.
When first invoked on a LAF-resource, it compiles it into efficient data structures and writes those to disk,
in such a way that this data can be loaded fast. This one-time compilation process takes roughly 15 minutes,
but then the data loads in a matter of seconds every time you want to work with it. Furthermore,
LAF-Fabric offers a programmers interface (API) to the LAF data,
by which the programmer can walk over the nodes and edges and collect feature information on the fly.
These walks are fast, and can be programmed easily.

The idea to create LAF-Fabric arose after we tried to use a library called graf-python \cite{Bouda2013},
part of POIO \cite{Bouda2012}, for the biblical LAF data.
Unfortunately, the way graf-python was programmed made it unsuitable for dealing with our LAF resource because of its size.
Python is a scripting language with a clean syntax and a good performance if used judiciously,
hence we undertook to write LAF-Fabric in Python as well.
We use those parts of Python that perform best for the heavy data lifting,
and those parts that are most user friendly for the programmers interface.
LAF-Fabric is a package that can be imported in any Python script,
and it behaves particularly well when invoked in an IPython notebook.

IPython Notebook is an interactive way of writing Python scripts and documentation \cite{Perez2007}.
A notebook is a document in which the data analyst writes cells with Python code and other cells with documentation.
Code cells can be run individually, in any order, while the results of the execution remain in memory.
The notebook has powerful capabilities of formatting results. A notebook can be published easily on the web,
so that others can download it and execute it as well, provided they have the same data and packages installed.
IPython notebook belongs to a branch of computer programming called scientific computing.
It is about explorative data analysis by means of computing power.
The scientific programmer produces analyses, charts, and documents that account for his data and results.
By contrast, the typical software engineer produces applications that perform well-defined tasks for end users.
The scientific programmer works close to the researchers, and writes special purpose code fast,
and reacts to changing demands in an agile way.
The software engineer works at a greater distance from the actual use cases.
He uses programming languages that support good software organization at the cost of a much slower development process.
He is less prepared to accomodate fast-changing requirements.

When LAF-Fabric runs in an IPython notebook, even the few seconds it needs to load data are required only once.
The programmer can experiment with his code cells at will, without the need to reload the data all the time.

LAF-Fabric has already been used for some significant data extractions.
There is a varied and growing set of notebooks \cite{Roorda2014b} on Github that is testimony to
the extent of use cases that can be served. Not only data analysis, but also adding new annotations is supported.
One of the use cases is the query saver itself.

\subsection{SHEBANQ}
The actual goal of the SHEBANQ project was to create a demonstrator query saver for the
ETCBC data. This has been achieved, and the resulting web application is called SHEBANQ \cite{Peursen2014b}.

\begin{figure}[htb]
\includegraphics[scale=0.36]{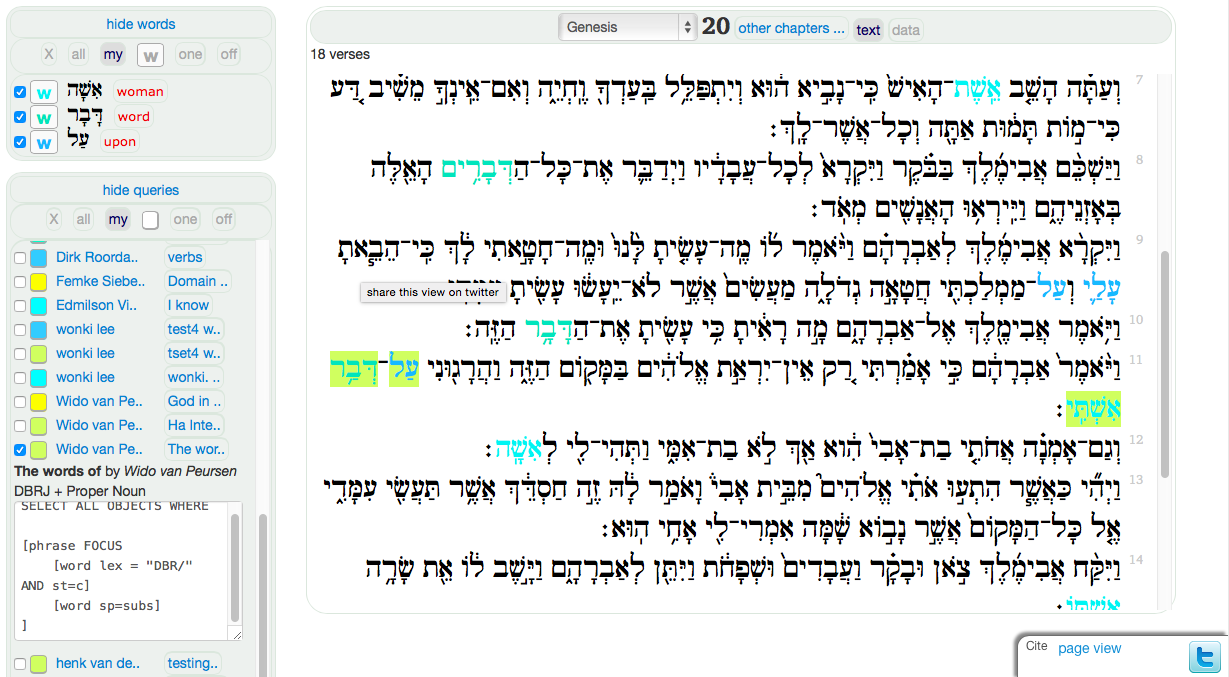}
\caption{A saved query in SHEBANQ}
\label{wordsof}
\end{figure}

\begin{figure}[htb]
\includegraphics[scale=0.36]{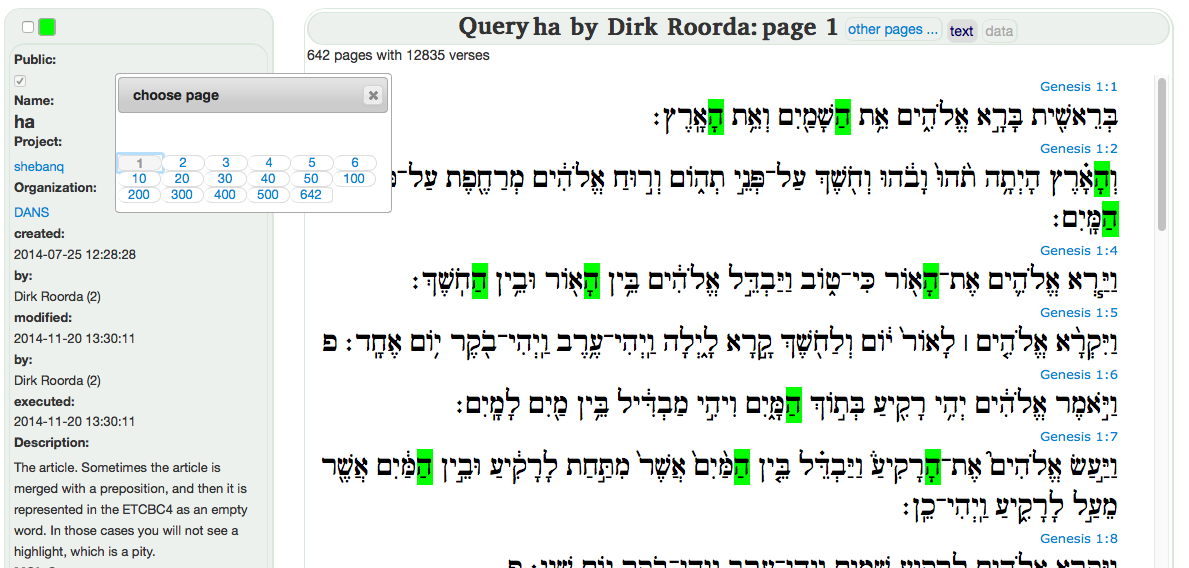}
\caption{Navigating through results of the query for the definite article (12,835 result verses)}
\label{qresults}
\end{figure}

It went live on 2014-08-01, and contains now, on 2015-01-06, 309 public queries, saved by 42 users.
The public part of the application offers users the options to read the Hebrew Bible chapter by chapter,
to see query results of public queries as annotations in the margin,
and to jump from query annotations to query descriptions and result lists.
Figure~\ref{wordsof} shows a screenshot of the page of a saved query.
No matter how many query results there are, the user is able so navigate through them all,
as can be seen in Figure~\ref{qresults}.

\begin{figure}[htb]
\includegraphics[scale=0.36]{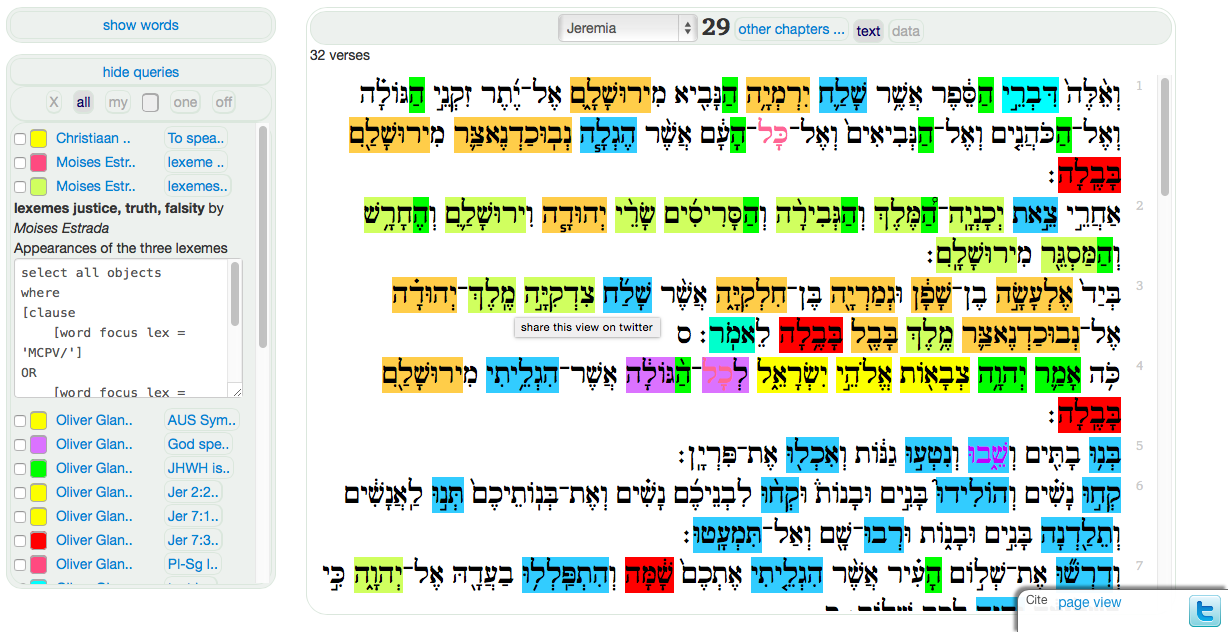}
\caption{Reading a passage and seeing the results of various queries.}
\label{qlist}
\end{figure}

When a user clicks on the verse indicator of a result, he is led to the browsing interface,
where other queries show up and can be navigated to, see Figure~\ref{qlist}.

When users register and log in, they can write their own queries, have them executed, save them,
including the query results, and make them public. In order to execute MQL queries,
SHEBANQ communicates with a web-service that is wrapped around the Emdros text database.

\begin{figure}[htb]
\includegraphics[scale=0.36]{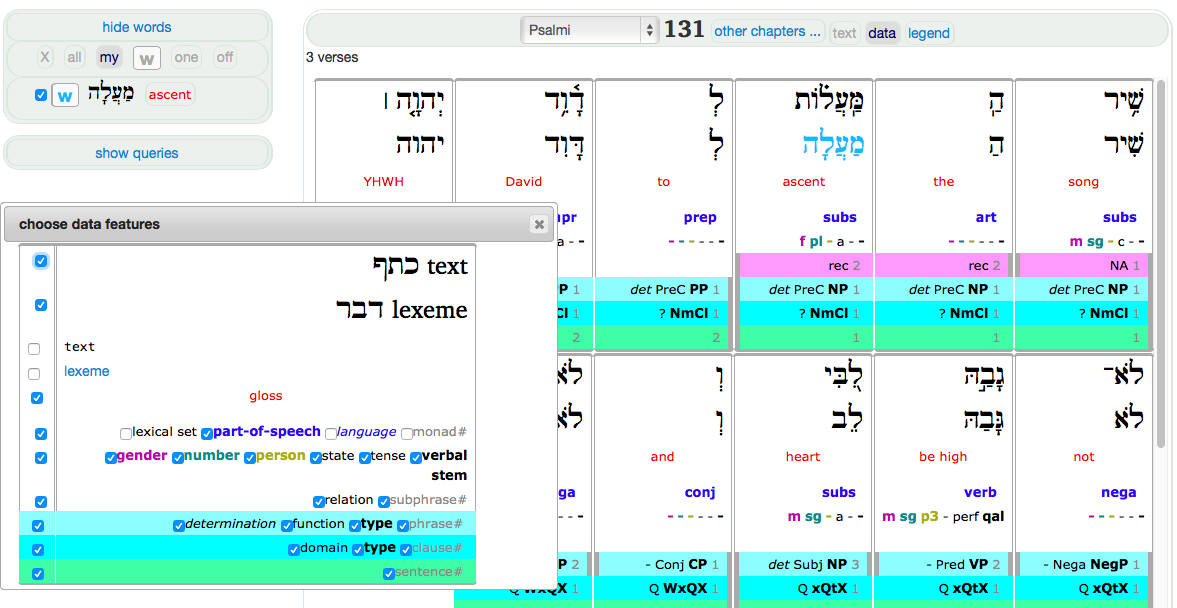}
\caption{Text and underlying data}
\label{textdata}
\end{figure}

While the underlying idea of SHEBANQ is straightforward, turning it into practice posed several challenges.
To begin with, the data had to be modeled in a way suitable for driving web applications.
We have programmed a MySQL export in a notebook, invoking LAF-Fabric.
Every now and then the functionality of SHEBANQ is extended.
For example, it can now show additional linguistic information in layers below the plain text.
We first experimented in LAF-Fabric by generating HTML for a visual prototype,
then we collectd feedback, and adapted our notebook.
We ran consistency checks whenever we wanted to make use of perceived regularities in the data.
After everything had crystallized out satisfactorily,
we built the new data representation into SHEBANQ, see Figure~\ref{textdata}.
The fact that both the notebook and the SHEBANQ website are written in Python turned out very convenient.

Rendering the Hebrew text turned out to be a problem because of subtle bugs in some platform/browser combinations.
On Mac OSX, Chrome and Safari mangled whitespace between particular combinations of consonants and diacritics.
We have explored the problems using LAF-Fabric and found a mitigating work-around.

If the demonstrator shows one thing, then it is the fact that there are many additional desiderata.
Whereas SHEBANQ has been designed to fulfill the sharing function, researchers also want to use it as a research tool.
It is not easy to write a good MQL query, because many of the linguistic aspects of the data are not shown on the interface.
If, for instance, a user wants to use the dictionary entries of the words or the syntactic features of clauses and phrases,
he has no immediate, visual clues.
So SHEBANQ has been extended again.
The user can now click on any word in the text for easy access to lexical information.

Other users see SHEBANQ as a preprocessor tool: they need data exports of query results.
The next iteration of SHEBANQ is planned to deliver that.

Another matter is usability: the number of queries is becoming too large to display them all in the margin.
Users want to be able to filter the queries they see on the basis of who wrote them,
both in the browsing interface and in the list of public queries.
Last-but-not-least, query execution is CPU-hungry.
We have already started thinking about measures to prevent excessive processor loads,
or ways to distribute the load over multiple servers.

\section{Reflection}
Back in 2012 we faced the challenge to provide better data models and better programs for biblical scholars.
It had become clear that the software companies that were developing the bible study applications
were not interested in building software for researchers. The researchers did not have funds to hire programmers themselves.
There seemed to be only one way out:
researchers should take their fate in their own hands and write the software themselves,
which looked like a daunting proposition at best and an impossible one at worst.
Yet, now in 2014, we have a publicly accessible tool for querying the linguistic data of the Hebrew Bible,
with a means to share those queries.
We also have a data laboratory where the programming theologian can take control over her data.
Collectively, biblical scholars can use the data laboratory to help the query tool evolve according to their needs.
Several factors have contributed to this achievement.

\begin{enumerate}
\item The existence of the LAF standard, which turned out to be a natural fit for this kind of data.
\item The realization that the plain text of the Hebrew Bible is not subject to copyright,
and hence that the ETCBC database of text and annotations can be made available as Open Source.
\item The existence of a research archive, DANS, acting as a data-hub;
the intellectual heritage of many years of ETCBC work lays deposited there and is open to scrutiny by anyone at any time.
\item The existence of a social medium for program code, Github;
all software for LAF-Fabric and SHEBANQ (and even some of the supporting software) lies there ready to be cloned and re-used.
\item The rise of scientific computing and its paraphernalia, such as (interactive) Python and auxiliary packages;
it offers an unprecedented level of user-friendliness to novice programmers;
it has the potential to draw a much wider range of humanities scholars into the enticing world of computing.
A researcher is much closer to a scientific programmer than to a software engineer.
\end{enumerate}

Yet, this is not sufficient to get the job done.
The ETCBC is steeped in its own ways, it has an efficient internal data workflow,
run with the best tools that were available in the late 1980s.
The internet existed then, but had not yet morphed into the world-wide web.
Data sharing is not in the genes of the ETCBC.
Doing unique things in relative isolation for a prolonged stretch of time tends to make you idiosyncratic.
The ETCBC has its own transliteration of Hebrew, its own,
locally documented way of coding data into forms that are optimal for local data processing. 

Opening up to the world poses new requirements on the ways the data is coded and how it is documented.
While we have archived the existing ETCBC documentation at DANS,
we started publishing a new kind of feature documentation on the web \cite{Roorda2014c}.
There we document not only the intended meaning of features,
but we also provide frequency lists of their complete value sets,
things that are easily computed by means of LAF-Fabric.

Can we say that we have succeeded in meeting the challenges posed in 2012?
It is too early for that. Proof of success would be the adoption of LAF-Fabric by at least some theological researchers,
interest in the Hebrew data from the side of computational linguistics and artificial intelligence,
and large access log files of the SHEBANQ web application.

At the moment of writing, all these indicators are non-zero
\cite{Roorda2014a}, \citename{Kalkman2013}~\citeyear{Kalkman2013,Kalkman2015},
which is promising, given the fact that we just started.

\section*{Acknowledgements}
The following people contributed significantly to the work described in this paper, in very different ways:

My colleagues at DANS: Henk Harmsen and Andrea Scharnhorst for granting additional time for research in these topics;
Henk van den Berg and Heleen van de Schraaf for developing fundamental portions of the web-service and web application
of SHEBANQ.

The ETCBC people: Eep Talstra and Constantijn Sikkel for creating and maintaining the ETCBC database
and sharing so much knowledge about it; Wido van Peursen for paving the way for increased sharing of data.
Oliver Glanz for showing order in the forest of features of the database;
Gino Kalkman and Martijn Naaijer for using the new tools and challenging me.
Grietje en Johan Commelin for their efforts to get LAF-Fabric working in the cloud.
Reinoud Oosting for finding subtle bugs in SHEBANQ and Janet Dyk introducing new use cases.

Wider Digital Humanities: Joris van Zundert (HuygensING)
for leading an inspiring Interedition bootcamp \cite{Zundert2012}
which set me on the track of rapid development for the humanities;
Rens Bod (Univ. of Amsterdam) and Andreas van Cranenburgh (Univ. of Amsterdam and HuygensING)
who asked for the Hebrew data as tree structures in order to try out Data Oriented Parsing for classical Hebrew.
\bibliographystyle{clin} 
\bibliography{hebrew-bible-data}  

\end{document}